\documentclass[11pt,a4paper]{article}
\usepackage[hyperref]{acl2019}
\usepackage{times}
\usepackage{latexsym}

\usepackage{url}

\usepackage{multirow}
\aclfinalcopy

\usepackage{bm}
\usepackage{CJKutf8}

\usepackage{enumitem}
\usepackage{amsfonts}
\usepackage{amssymb}

\usepackage{amsmath}
\usepackage{graphicx}
\usepackage{color}
\usepackage{subfigure}
\usepackage{booktabs}
\usepackage{float}

\newcommand*{\email}[1]{\texttt{#1}}

\setitemize[1]{itemsep=5pt,partopsep=1pt,parsep=\parskip,topsep=3pt}

\title{PKUSEG: A Toolkit for Multi-Domain \\ Chinese Word Segmentation}
\author{
Ruixuan Luo,\textsuperscript{\rm 2} \ 
Jingjing Xu,\textsuperscript{\rm 1} \ 
Yi Zhang,\textsuperscript{\rm 1} \ 
Zhiyuan Zhang,\textsuperscript{\rm 1} \
Xuancheng Ren,\textsuperscript{\rm 1} \ 
Xu Sun \textsuperscript{\rm 1,2}
\\
\textsuperscript{\rm 1}MOE Key Lab of Computational Linguistics, School of Computer Science, Peking University  \\
\textsuperscript{\rm 2}Center for Data Science, Peking University\\
\email{\{luoruixuan97,jingjingxu,zhangyi16,zzy1210,renxc,xusun\}@pku.edu.cn }
}

\begin{document}

\maketitle

\begin{abstract}
    Chinese word segmentation (CWS) is a fundamental step of Chinese natural language processing. In this paper, we build a new toolkit, named \textbf{PKUSEG}, for multi-domain word segmentation. Unlike existing single-model toolkits, PKUSEG targets multi-domain word segmentation and provides separate models for different domains, such as web, medicine, and tourism. Besides, due to the lack of labeled data in many domains, we propose a domain adaptation paradigm to introduce cross-domain semantic knowledge via a translation system. Through this method, we generate synthetic data using a large amount of unlabeled data in the target domain and then obtain a word segmentation model for the target domain. We also further refine the performance of the default model with the help of synthetic data. Experiments show that PKUSEG achieves high performance on multiple domains. The new toolkit also supports POS tagging and model training to adapt to various application scenarios. The toolkit is now freely and publicly available for the usage of research and industry.\footnote{\url{https://github.com/lancopku/pkuseg-python/}}
\end{abstract}

\section{Introduction}
Chinese word segmentation is a fundamental task of Chinese processing.  Since words define the basic semantic unit of Chinese, the quality of segmentation directly influences the performance of downstream tasks. 
In recent years, Chinese word segmentation has undergone great development. The best-performing systems are mostly based on conditional random fields (CRF)~\cite{Lafferty01conditionalrandom,DBLP:conf/acl/SunWL12}. However, despite the promising results, these approaches heavily rely on feature engineering. To tackle this problem, many researchers \cite{DBLP:conf/emnlp/ChenQZLH15,DBLP:conf/acl/CaiZ16,DBLP:conf/ijcai/LiuCGQL16,xu2016dependency} explore neural networks to automatically learn better representations. 

Recently, there arise several public segmentation toolkits, such as jieba, HanLP, and so on. For efficiency, they are built on traditional segmentation models, like perceptron~\cite{conf/acl/ZhangC07} or CRF, rather than time-consuming neural networks.  These toolkits only provide a single coarse-grained segmentation model, mostly trained on news domain data. In real-world applications, the domain of text varies and the text from different domains has different domain-specific segmentation rules. This increases the difficulty of segmentation and drops the performance of existing toolkits on text from various domains.

%
To address this challenge, we propose a multi-domain segmentation toolkit, PKUSEG, based on the work of \citet{DBLP:conf/acl/SunWL12}: We adopt a fast and high-precision model CRF as implementation. PKUSEG includes multiple pre-trained domain-specific segmentation models. Since some domains may be of low resources, we use a pre-training technique to improve the quality of segmentation. We first pre-train a coarse-grained model on a mixed corpus, including millions of data from news and web domains. Then, we fine-tune the coarse-grained model on specific domain data to get fine-grained models. 

To train a model for some domains that do not have any manually annotated data, but have a lot of unlabeled data, we further propose an unsupervised domain adaptation approach named masked Attention Augmentation. Due to the lack of labeled data in the target domain, these domains lack the semantic knowledge of the target domain, so it is difficult to correctly segment the proper noun or syntactic structure of the target domain. To address these issues, we propose to learn word-level semantic knowledge of the target domain by introducing translation. We first obtain the English translation of the data through a translation model. Then, we train a large Transformer model for segmentation and masked language model tasks. The model takes the translation pairs as input and tries to obtain the semantic knowledge of the target domain by learning the word-level alignment between the two languages. The Transformer model is then used to generate synthetic data for the target domain. We use synthetic data to further train the target domain model or improve the general model performance.

In addition to provided segmentation models, PKUSEG also allows users to train a new model on their domain data.  Furthermore, POS tagging is supported in PKUSEG to adapt to various scenarios. Experimental results show that PKUSEG achieves high performance on multi-domain datasets. 

In summary, PKUSEG has the following characteristics:
\begin{itemize}
    \item \textbf{Good out-of-the-box performance.} The default word segmentation model provided by PKUSEG is trained on a large-scale, curated, multi-domain dataset, which shows stable and high performance across various domains.
    \item \textbf{Domain-specific pre-trained models.} PKUSEG also comes with multiple pre-trained models that are fine-tuned on texts of different domains, which further elevates domain-specific performance, suitable for analyzing in-domain texts. 
    \item \textbf{Easy transfer learning.} For advanced users, PKUSEG supports transfer learning based on the default multi-domain model. Users can fine-tune the model on their custom segmented texts.
    \item \textbf{Unsupervised domain adaptation.} For the domains which do not have any manually annotated data, but have a lot of unlabeled data, we propose an unsupervised domain adaptation paradigm. And our method successfully trains the domain-specific model without any target domain annotated data.
    \item \textbf{POS tagging.} PKUSEG also provides users with POS tagging interfaces for further lexical analysis.
\end{itemize}

\section{ Implementation}
This section gives a detailed description of toolkit implementation.

\subsection{Conditional Random Field}

Despite the better performance, we do not use neural networks as an implementation due to the time-consuming training process. Instead, we use a well-performing and fast-training model, CRF, as implementation, considering the trade-off between time cost and high accuracy.  We optimize the weights of CRF by maximizing the log-likelihood of the tags of the reference sequence. When calculating the log-likelihood, the log-likelihood function can be calculated by the recursive algorithm in linear time. During inference, the Viterbi~\cite{Forney:1973ly} algorithm is adopted. The goal is to find the sequence of tags by dynamic programming.

\subsection{ADF Algorithm}

For CRF with many high-dimensional features, the amount of parameters is very large, leading to expensive training costs. To address this problem, we use adaptive online gradient descent based on feature frequency information (ADF)~\cite{DBLP:conf/acl/SunWL12} for training.
The ADF algorithm does not use a single learning rate for all parameters like stochastic gradient descent (SGD), instead, it turns the learning rate into a vector with the same dimension as the parameters. The learning rate of each parameter is automatically adjusted according to the frequency of the parameter. The idea is that the feature with higher frequency will be more adequate. 

\begin{figure*}[t]
    \centering
    \includegraphics[scale=0.35]{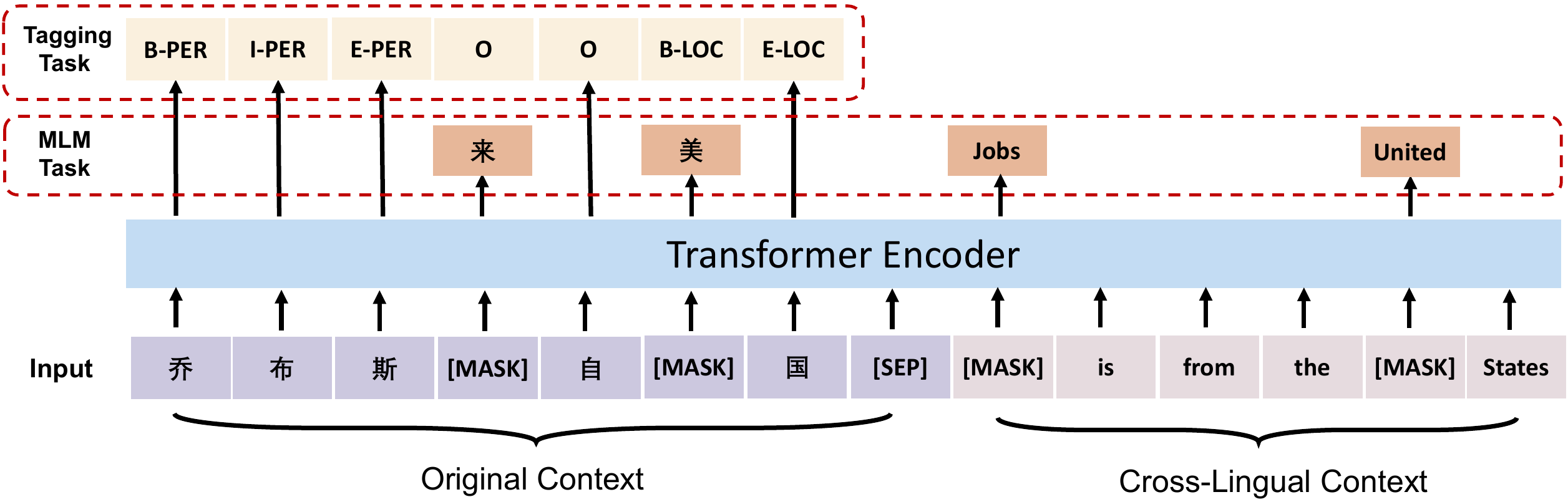}
    \caption{An illustration of the proposed approach $m\mathcal{A}^2$. The tagging task is only applied to the original context while the masked language modeling is performed on both contexts. \protect\footnotemark}
    \label{fig:model}
\end{figure*}

\subsection{Pre-training}
To handle the problem of low-resource for some domains, we adopt a pre-training technique in PKUSEG following the work of~\citet{journals/corr/XuS17}. We mix news and web data together as pre-training data. News data comes from dataset PKU provided by the Second International Chinese Word Segmentation Bakeoff. Web data comes from the Weibo dataset provided by NLPCC-ICCPOL 2016 Shared Task~\cite{conf/nlpcc/QiuQS16}. A hybrid dataset CTB  is also involved in pre-training. In the process of fine-tuning,  models are initialized with the pre-trained model and trained on domain-specific data.  So far PKUSEG supports four fine-grained domains, including news, medicine, tourism, and web.  Considering the covered domains are limited, we also provide a pre-trained model for generalization.

\subsection{A Large-Scale Vocabulary}
One major difficulty of multi-domain segmentation is spare domain-specific words. It is hard to cover all of these words on the training set. Therefore,  to increase the coverage rate of PKUSEG, we automatically build a large-scale domain vocabulary. The word resource is crawled from Sogou website\footnote{\url{https://pinyin.sogou.com/dict/}} and extracted from the training data of PKU, MSRA, Weibo, and CTB.  In total, we extract almost 850K words.  The distribution of words is shown in Table~\ref{tab:dist}.

\begin{table}[h]
    \centering
    \begin{tabular}{c|c}
    \hline
          Domain  & Vocabulary Size  \\
         \hline
          Medicine & 447K\\
          Location & 117K \\
          Name & 105K \\
          Idiom & 50K \\
         Organization & 31K \\
         Training Words & 100K\\
         \hline
          total & 850K\\

    \hline
    \end{tabular}
    \caption{The distribution of words in the extracted vocabulary.}
    \label{tab:dist}
\end{table}

\subsection{Derive Synthetic Data with masked Attention Augmentation}

In the process of practical application, because the cost of data annotation is very expensive, the domain where the model is deployed often faces a situation where unlabeled samples are abundant but labeled samples are seriously insufficient. In order to make full use of the existing labeled data and unlabeled data of the target domain to train a domain-specific model, we propose a masked Attention Augmentation ($m\mathcal{A}^2$) approach \cite{DBLP:conf/emnlp/LuoZCS21} to derive synthetic data. Our method is able to train a segmentation model for the target domain without any target domain labeled data. 

An overview of $m\mathcal{A}^2$ is illustrated in Figure~\ref{fig:model}. Specifically, we first use Google Translate to obtain an English translation of the source domain and target domain data. Then we train a Transformer model of the target domain. Given a translation pair, we randomly mask some tokens and encourage the model to reconstruct them. Since the tokens of both languages can be masked, this masked language model task encourages the model to recover the tokens based on the prompts in another language, and further enhance the alignment between the two languages. Meanwhile, we train the word segmentation task on the source domain. The final objective is the sum of the masked language model task and the Chinese word segmentation task. The $m\mathcal{A}^2$ approach prompts the model to learn about the crucial word-level semantic knowledge of the target domain. And then, the well-trained Transformer model is used to derive synthetic data from the target domain. The synthetic data is finally used for training domain-specific models or improving the generalization performance of the default model. So far, we obtain unlabeled data of three domains from Wikipedia, including science, art, and entertainment. And we use the news domain as our source domain.

\footnotetext{This figure is from \cite{DBLP:conf/emnlp/LuoZCS21}.}
 
\section{Usage}
PKUSEG has high precision performance along with user-friendly interfaces. It is developed based on standard python3  libraries. PKUSEG supports common running platforms,  such as Windows, Linux, and MacOS. 

\subsection{Installation}
 PKUSEG offers two user-friendly installation methods. Users can easily install it with PyPI and the corresponding models will be downloaded at the same time. A typical command is:
\begin{center}
\fbox{
pip3 install pkuseg
}
\end{center}
Users also can install PKUSEG from GitHub. After downloading the project code from GitHub, users can run the following command to install PKUSEG:
\begin{center}
\fbox{
python setup.py build\_ext -i
}
\end{center}
Noting that the downloaded project from GitHub does not include pre-trained models, users need to additionally download them from GitHub or train a new model.
\subsection{Segmentation}
The followings are a detailed introduction to segmentation interfaces.

\begin{figure}[h]
    \centering
    \includegraphics[width=0.95\linewidth]{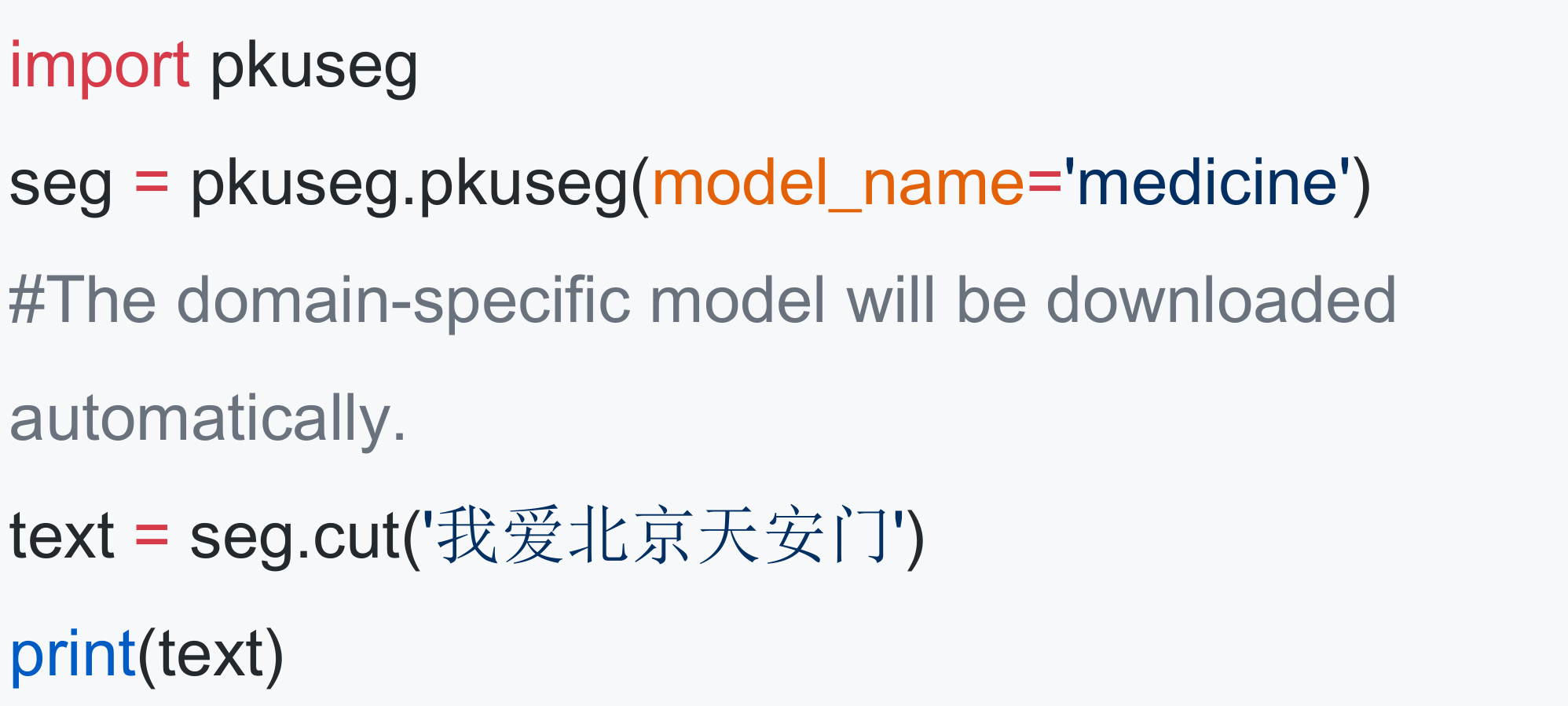}
    \caption{An example code of specifying the model of ``medicine'' domain.}
    \label{fig:demo_1}
\end{figure}

\paragraph{Domain-specific Segmentation.}
 If a  user is aware of the domain of the text to be segmented, then he/she can use the domain-specific model. An example code for specifying a model is shown in Figure~\ref{fig:demo_1}. If a model is toolkit-provided,  users can directly use the domain name to call it, e.g, ``medicine'', ``tourism'', ``web'', and ``news''. The model is automatically loaded based on the parameter ``model\_name''. If the model is user-trained, ``model\_name'' refers to the model path.

\begin{figure}[h]
    \centering
    \includegraphics[width=0.95\linewidth]{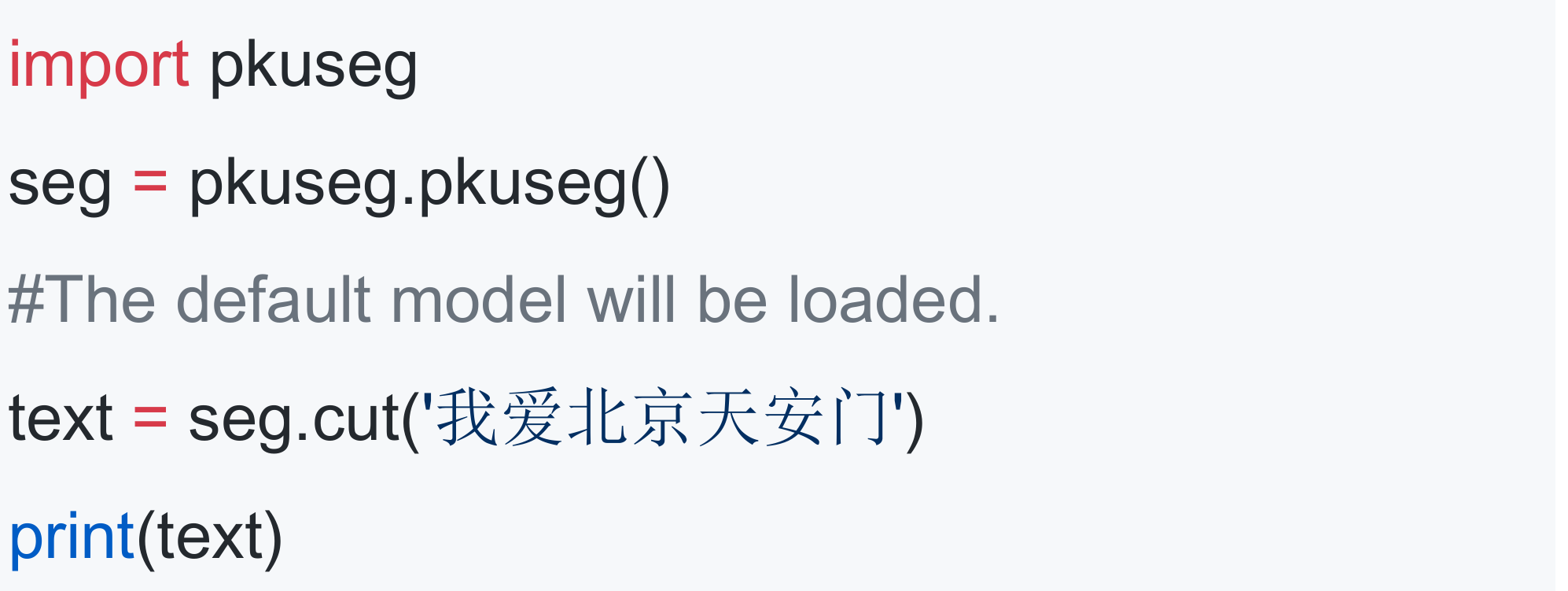}
    \caption{An example code of segmentation with the default model.}
    \label{fig:demo_2}
\end{figure}

\paragraph{Coarse-grained Segmentation.}
Although PKUSEG is designed to satisfy the situation where users know the domain of the text to be segmented, we also provide a coarse-grained model in case that the user can not distinguish the target domain. The coarse-grained model works under the default mode.  Figure~\ref{fig:demo_2} shows an example code using the default mode. 

\begin{figure}[h]
    \centering
    \includegraphics[width=0.95\linewidth]{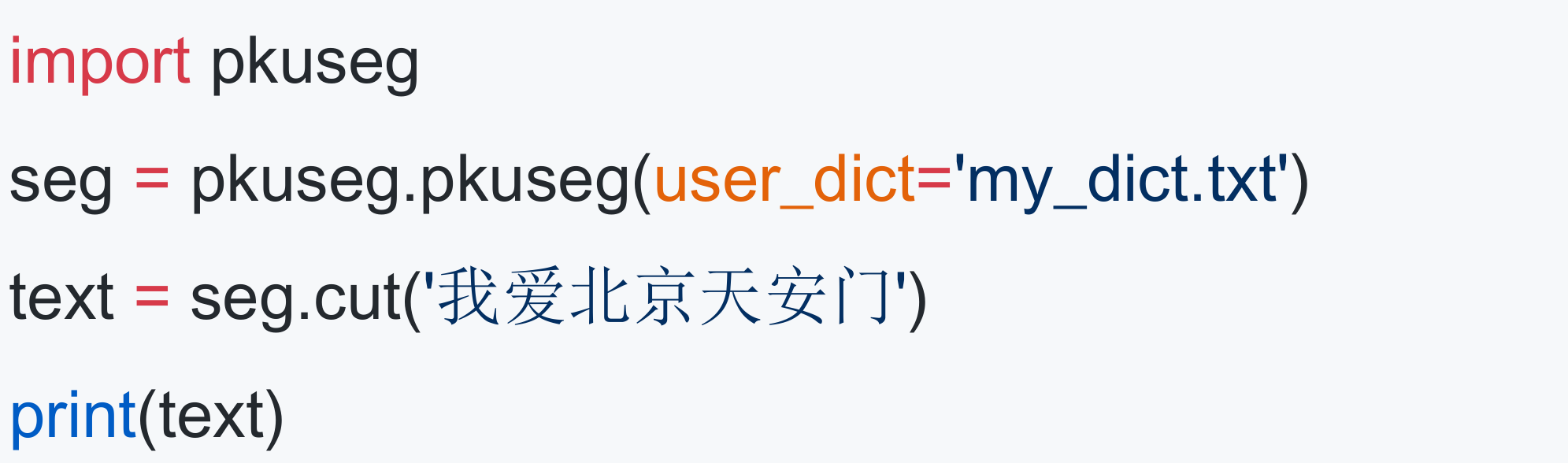}
    \caption{An example code of using a user-defined dictionary.}
    \label{fig:demo_3}
\end{figure}

\paragraph{User-defined Dictionary.}
To better recognize new words, users can add a dictionary to cover the words that do not occur in the dictionary of PKUSEG. The provided dictionary file should follow the following format. Each row has a single word and the dictionary file is encoded in the UTF-8 format. Figure~\ref{fig:demo_3} shows the usage of a user-defined dictionary.

\begin{figure}[h]
    \centering
    \includegraphics[width=0.95\linewidth]{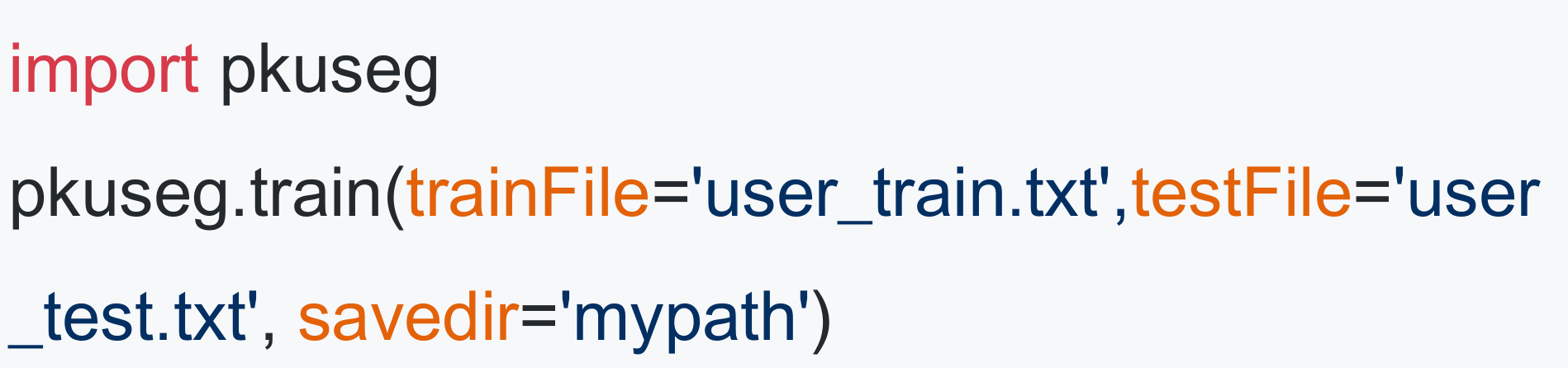}
    \caption{An example code of training a new model with user-provided data.}
    \label{fig:demo_5}
\end{figure}

\paragraph{Model Training.}
PKUSEG also allows users to train a new model from scratch with their training data. Figure~\ref{fig:demo_5} is an example code for showing how to train a new model. 

\begin{figure}[h]
    \centering
    \includegraphics[width=0.95\linewidth]{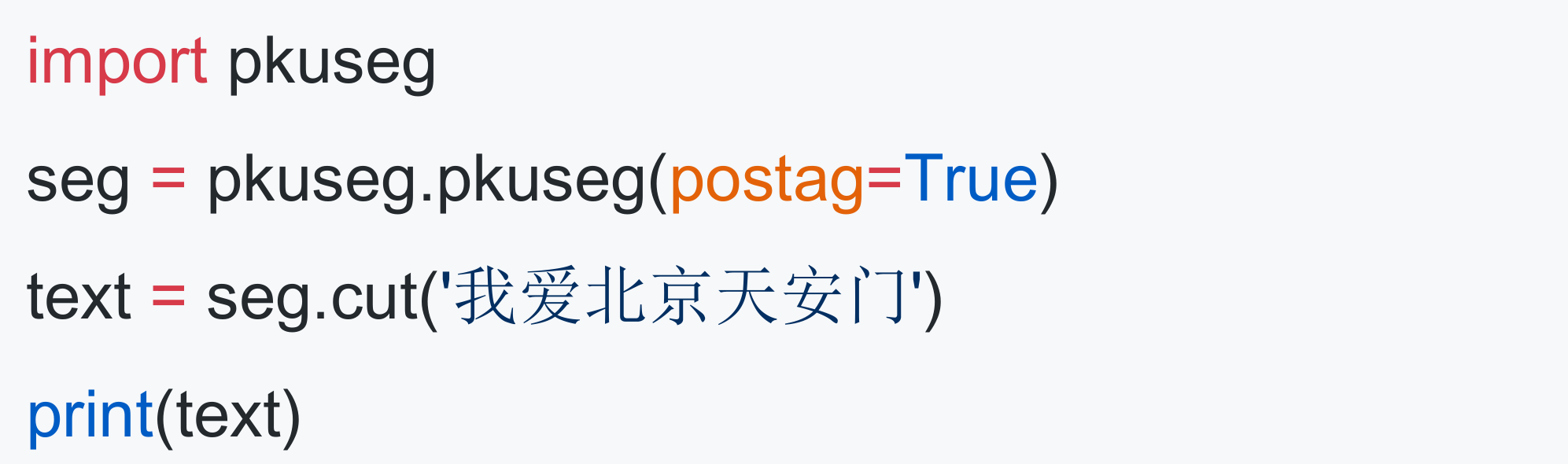}
    \caption{An example code of segmenting and POS tagging.}
    \label{fig:demo_4}
\end{figure}

\paragraph{Segmentation with POS Tagging.} 
In addition to segmentation, PKUSEG  can also label POS tags for words in a sentence. The usage of POS tagging interfaces is shown in Figure~\ref{fig:demo_4}.

\section{Experiment}
This section evaluates the performance of PKUSEG.

\subsection{Dataset}
\paragraph{MSRA \& PKU.} MSRA and PKU are from the news domain and provided by the Second International Chinese Word Segmentation Bakeoff. 

\paragraph{CTB8.} Chinese Tree Bank is a hybrid domain dataset.\footnote{\url{https://catalog.ldc.upenn.edu/LDC2013T21}} It consists of approximately 1.5 million words from Chinese newswire, government documents, magazine articles, various broadcast news, and broadcast conversation programs, web newsgroups, and weblogs.

\paragraph{Weibo.} This dataset comes from the NLPCC-ICCPOL 2016 Shared Task. Different from the popularly used newswire datasets, this dataset consists of many informal micro-texts.

\paragraph{Medicine \& News \& Tourism.} The corpus
is originally constructed in~\citet{LiqunQiu} by
annotating multi-domain texts. 

\paragraph{Science \& Art \& Entertainment.} To evaluate the performance of domain-specific models constructed by our unsupervised method, we manually annotated about 100 sentences for each of the three domains on Wikipedia.

          

\subsection{Out-of-domain Results}
To show the effect of domain knowledge on segmentation performance, we train a model on the CTB8 dataset and report its performance on different datasets. Here we choose CTB8 as an example because CTB8 is a hybrid dataset. The results are shown in Table~\ref{tab:outofodmain}. 

\begin{table}[H]
    \centering
    \begin{tabular}{l|c|c}
    \hline
       CTB8 Training   & Testing  &  F1  \\
         \hline
         In-domain & CTB8  & 	95.69 \\
          \hline
       \multirow{3}{*}{Out-of-domain}  & MSRA (News)	  & 83.67\\
          
          & PKU (News)  & 	89.67\\
         & Weibo (Web) & 	91.19 \\\hline
      \multirow{2}{*}{Average}    & All Average & 90.06\\
          & OOD Average & 	88.18\\
    \hline
    \end{tabular}
    \caption{Results of PKUSEG with a model trained on CTB.  ``All Average'' is the average F1 score of all datasets. ``OOD Average (Out-of-domain Average)'' is the average results of datsets except for CTB.}
    \label{tab:outofodmain}
\end{table}

Here are the results without using vocabulary. We can see that the performance drops obviously on out-of-domain datasets. Since different domains have their unique segmentation standard, it is not suitable to provide one single model for various domain data. The result demonstrates the necessity of fine-grained segmentation toolkits.

\subsection{Pre-training Results}
We combine existing large-scale datasets together, including  PKU (news), Weibo (web), and CTB8 (hybrid), and use them as pre-training data to obtain a coarse-grained model. Then the coarse-grained model is used to fine-tune domain-specific models.

\begin{table}[H]
    \centering
    \begin{tabular}{c|c|c}
    \hline
    
            & w/o Pre-train & w. Pre-train  \\
         \hline
          Medicine  &95.61 &95.10 \\
          Web & 94.75 & 95.49 \\
          News  &97.58	&97.80 \\
          Tourism  &96.36	&97.10 \\ \hline
          Average & 96.08 & 96.37 \\
    \hline
    \end{tabular}
    \caption{ F1 scores of PKUSEG with and without pre-training. Here are results without using  dictionaries. Web data comes from the Weibo dataset.}
    \label{tab:pre-train}
\end{table}

Table~\ref{tab:pre-train} shows the effect of pre-training. The pre-training models perform much better in terms of average score, especially on datasets with lower resource (e.g., tourism). 



         

\subsection{Default Performance}
Considering the fact that many users tend to use the default mode to test performance,  with the default model and vocabulary of PKUSEG. We also report experimental results on the default mode. The results are shown in Table~\ref{tab:default}.

\begin{table}[H]
    \centering

    \begin{tabular}{c|c}
         \hline
          PKUSEG &  F1 \\
         \hline
         MSRA & 87.29 \\
         CTB8 & 91.77\\
         PKU & 93.43\\
         Weibo & 92.68\\
         
         \hline
    \end{tabular}
    \caption{Performance of PKUSEG on different domains with the default mode. }
    \label{tab:default}
\end{table}

As we can see, the performance of the default model performs worse than that of domain-specific models. Therefore, we recommend users use domain-specific models, rather than the default model if the user can distinguish the domain of the text. 

To learn more about the practical application of PKUSEG, we also show some segmentation examples that randomly crawled from articles that cover the domains of medicine, travel, web text, and news. The segmentation results are shown in Table~\ref{tab:cases}. PKUSEG has high accuracy when dealing with words that need professional domain knowledge.

\begin{CJK*}{UTF8}{gbsn}
\begin{table}[!t]
    \centering
    \small
    \begin{tabular}{p{7cm}}
    \hline
    \hline
    \multicolumn{1}{c}{Medicine}\\
    \hline
    \hline
    
    医联~ 平台~ ： 包括~ 挂号~ 预约~ 查看~ 院内~ 信息~ 化验单~ 等~ ，~ 目前~ 出现~ 与~ 微信~ 、 ~支付宝~ 结合的~ 趋势~ 。\\
    Medical Association platform includes registration appointment,  in-hospital information management, etc. There is a trend of integration with WeChat and Alipay.\\
    
    \hline
    \hline
    \multicolumn{1}{c}{Travel} \\
    \hline
    \hline
    在 ~这里~ 可以 ~俯瞰~~~ 维多利亚港~~ 的~ 香港岛~ ， ~九龙 ~半岛~ 两岸 ~，~ 美景~ 无敌 。\\
    It overlooks Victoria Harbour and the two sides of the Kowloon Peninsula. The view is so beautiful. \\ 
    \hline
    \hline
    \multicolumn{1}{c}{Web} \\
    \hline
    \hline
    
    【~ 这是~ 我 ~的 ~世界 ~， ~你 ~还 ~未~ 见~ 过~ 】~ 欢迎~ 来 ~参加 ~我~ 的 ~演唱会~ 听点~ 音乐\\
This is my world that you have not seen before. Welcome to participate in my concert to listen to music. \\
    \hline
    \hline
    \multicolumn{1}{c}{News} \\
    \hline
    \hline
    他~ 不~ 忘~ 讽刺~ 加州~ ：~ “ ~加州~ 已~ 在~ 失控~ 的~ 高铁~ 项目~ 上~ 浪费~ 了~ 数十亿美元 ~， ~完全~ 没有~ 完成~ 的 ~希望~ 。 \\
     He did not forget to satirize California, ``California has been wasting billions of dollars on the uncontrolled high-speed rail projects, which is of no hope being completed at all''.\\
    乌克兰~ 政府 ~正式~ 通过~ 最新~ 《 ~宪法~ 修正案~ 》~ ， ~正式~ 确定~ 乌克兰~ 将~ 加入~ 北约~ 作为~ 重要 ~国家~ 方针~ ，~ 该~ 法~ 强调 ~，~ "~ 这项~ 法律~ 将~ 于 ~发布~ 次日~ 起 ~生效~ "~ 。\\
    The Ukrainian government officially adopted the latest Constitutional Amendment,  confirming that Ukraine will regard joining the NATO as an important national policy. The law emphasizes that it will take effect from the next day.\\
    \hline
    \hline
    \end{tabular}
    \caption{Examples of segmentation on various domains with the domain-specific models.}
    \label{tab:cases}
\end{table}
\end{CJK*}

\subsection{Unsupervised Domain-specific Performance}

To verify the practicability of domain-specific models constructed by our unsupervised approach, we conduct experiments with data from three domains of science, art, and entertainment, and compare the performance of our domain-specific models with the default PKUSEG model. The results are shown in Table \ref{tab:unsupervised-domain}. 

\begin{table}[H]
    \centering
    \scalebox{0.9}{
    \begin{tabular}{c|c|c}
    \hline
    
            & pkuseg(default) & pkuseg-domain  \\
         \hline
          Science  & 90.20 & 90.73 \\
          Art & 90.77 & 92.47 \\
          Entertainment  & 90.23 & 90.77 \\
          \hline
          Average & 90.40 & 91.32 \\
    \hline
    \end{tabular}}
    \caption{ Performance of the domain-specific model constructed by our unsupervised approach comparing with the default PKUSEG model.}
    \label{tab:unsupervised-domain}
\end{table}

The domain-specific models have an average +$0.92$ improvement in the three domains.

\begin{table}[H]
    \centering
    \begin{tabular}{c|c|c}
    \hline
    
            & pkuseg(default) & pkuseg-refine  \\
         \hline
          MSRA  & 87.29 & 89.49 \\
          CTB8 & 91.77 & 90.91 \\
          PKU  & 93.43 & 93.38 \\
          Weibo & 92.68 & 91.70 \\
          \hline
          Average & 91.29 & 91.37 \\
    \hline
    \end{tabular}
    \caption{ In-domain performance of the default PKUSEG model and refined PKUSEG model.}
    \label{tab:refine-id}
\end{table}

\subsection{Refined Default Performance}

To improve the generalization performance of the default model, we combine all the synthetic data and annotated data together, and then train the refined PKUSEG model with this large mixed dataset. We compare this refined model with the original default PKUSEG model on both in-domain datasets (the training sets of the original default model) and out-of-domain datasets (the unannotated domains from Wikipedia). The experimental results are shown in Table~\ref{tab:refine-id} and Table~ \ref{tab:refine-ood}.

The refined model has no performance drop in the in-domain datasets while having +$0.67$ F1 improvement on the out-of-domain datasets. The results show that the refined model achieves better generalization performance.

\begin{table}[H]
    \centering
    \scalebox{0.9}{
    \begin{tabular}{c|c|c}
    \hline
    
            & pkuseg(default) & pkuseg-refine  \\
         \hline
          Science  & 90.20 & 90.52 \\
          Art & 90.77 & 92.16 \\
          Entertainment  & 90.23 & 90.52 \\
          \hline
          Average & 90.40 & 91.07 \\
    \hline
    \end{tabular}}
    \caption{ Performance of the default PKUSEG model and refined PKUSEG model on specific domains.}
    \label{tab:refine-ood}
\end{table}

\section{Conclusion and Future Work}
In this paper, we propose a new toolkit PKUSEG for multi-domain Chinese word segmentation. PKUSEG provides simple and user-friendly interfaces for users. Experiments on widely-used datasets demonstrate that PKUSEG performs well with high accuracy.  So far PKUSEG supports domains like medicine, tourism, web, and news.  In the future, we plan to release more domain-specific models and improve the efficiency of PKUSEG further.

\bibliography{acl2019}
\bibliographystyle{acl_natbib}

\end{document}